%% file: arxiv.tex
\documentclass[10pt,twocolumn,letterpaper]{article}

\usepackage[pagenumbers]{arxiv} % To force page numbers, e.g. for an arXiv version

\input{preamble}

\definecolor{iccvblue}{rgb}{0.21,0.49,0.74}
\usepackage[pagebackref,breaklinks,colorlinks,allcolors=iccvblue]{hyperref}

\def\paperID{9442} % *** Enter the Paper ID here
\def\confName{ICCV}
\def\confYear{2025}
\newcommand{\Ours}{MaskCLIP++}
\newcommand{\myPara}[1]{\vspace{0.5em}\noindent\textbf{#1}}
\newcommand{\uline}[1]{\underline{#1}}
\definecolor{GrayBack}{HTML}{EFEFEF}
\newcommand{\cmark}{\ding{51}} % 对勾符号
\newcommand{\xmark}{\ding{55}} % 八叉符号
\definecolor{RedFore}{HTML}{C25759}
\definecolor{BlueFore}{HTML}{599CB4}
\newcommand{\redsub}[1]{\textcolor{RedFore}{\raisebox{-0.5ex}{\scriptsize #1}}}

\title{High-Quality Mask Tuning Matters for Open-Vocabulary Segmentation}

\author{
    Quan-Sheng Zeng$^1$
    \quad
    Yunheng Li$^1$
    \quad
    Daquan Zhou$^3$
    \quad
    Guanbin Li$^4$
    \quad
    Qibin Hou$^{1,2}$\thanks{Corresponding author.}
    \quad
    Ming-Ming Cheng$^{1,2}$ \\
    $^{1}$VCIP, School of Computer Science, Nankai University \quad
    $^{2}$NKIARI, Futian, Shenzhen \\
    $^{3}$ByteDance Inc.\quad
    $^{4}$School of Computer Science and Engineering, Sun Yat-sen University \\
    {\tt\small qszeng@mail.nankai.edu.cn}
    \quad
    {\tt\small \{houqb, cmm\}@nankai.edu.cn}
}

\begin{document}
\maketitle
\input{sec_arxiv/0_abstract}    
\input{sec_arxiv/1_intro}

\input{sec_arxiv/2_related}

\input{sec_arxiv/3_method}

\input{sec_arxiv/4_experiments}
\input{sec_arxiv/5_discussion}
{
    \small
    \bibliographystyle{ieeenat_fullname}
    \bibliography{arxiv}
}
\input{sec_arxiv/6_supplementary}
\end{document}

%% file: preamble.tex
\usepackage{amsmath,amsfonts,bm}
\usepackage{graphicx}
\usepackage{subcaption}
\usepackage{pifont}
\usepackage{multirow}
\usepackage{makecell}
\usepackage{tablefootnote}
\usepackage{booktabs}
\usepackage{tikz}
\usepackage{contour}

%% file: sec_arxiv/0_abstract.tex
\begin{abstract}
Open-vocabulary image segmentation has been advanced through the synergy between mask generators and vision-language models like Contrastive Language-Image Pre-training (CLIP).
Previous approaches focus on generating masks while aligning mask features with text embeddings during training.
In this paper, we observe that relying on generated low-quality masks can weaken the alignment of vision and language in regional representations.
This motivates us to present a new fine-tuning framework, named \Ours, which uses ground-truth masks instead of generated masks to enhance the mask classification capability of CLIP.
Due to the limited diversity of image segmentation datasets with mask annotations, we propose incorporating a consistency alignment principle during fine-tuning, 
which alleviates categorical bias toward the fine-tuning dataset.
After low-cost fine-tuning, \Ours{} significantly improves the mask classification performance on multi-domain datasets.
Combining with the mask generator in previous state-of-the-art mask-based open vocabulary segmentation methods,
we achieve performance improvements of +1.7, +2.3, +2.1, +3.1, and +0.3 mIoU on the A-847, PC-459, A-150, PC-59, and PAS-20 datasets, respectively.
Code is avaliable at \url{https://github.com/HVision-NKU/MaskCLIPpp}.
\end{abstract}

%% file: sec_arxiv/1_intro.tex
\section{Introduction}
\label{sec:intro}

\begin{figure}[tbp]
  \centering
  \begin{subfigure}[b]{0.49\textwidth}
      \centering
      \includegraphics[width=\textwidth]{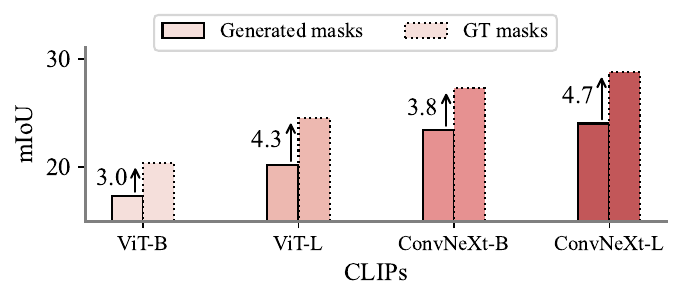}
      \caption{Performance: generated masks v.s. ground truth masks.}
      \label{subfig:oracle_mask_results}
  \end{subfigure}
  % \hfill
  \begin{subfigure}[b]{0.49\textwidth}
      \centering
      \includegraphics[width=\textwidth]{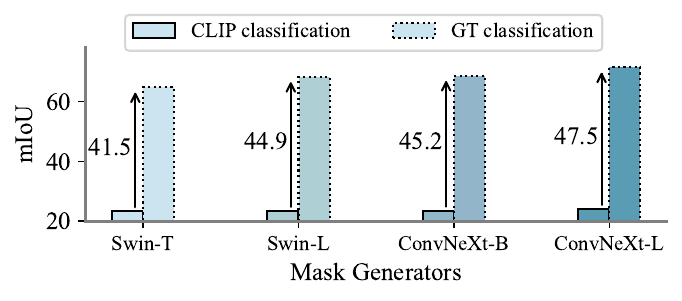}
      \caption{Performance: CLIP classification v.s. ground truth classification.}
      \label{subfig:oracle_cls_results}
  \end{subfigure}
  \vspace{-1.5em}
  \caption{
    Observations: \protect{\subref{subfig:oracle_mask_results}} demonstrates the potential negative impact of low-quality generated masks on CLIP's mask classification learning,
    while \protect{\subref{subfig:oracle_cls_results}} showcases the untapped generalization potential of existing mask generators.
    %  (a) Performance comparison of various original CLIP architectures using the same generated and ground truth masks.
    %  (b) Performance comparison of various mask generators with different backbones using CLIP based on ConvNeXt-L and ground truth classification.
    Mask generators are trained on the COCO~\cite{lin2014mscoco}, with results reported on the ADE20K~\cite{zhou2017ade20k}.
   }
   \vspace{-5pt}
  \label{fig:oracle_results}
\end{figure}

Image segmentation is one of the most extensively studied tasks in computer vision, 
which aims to partition an image into several regions where each region corresponds to the same object or shares consistent semantics.
Traditional image segmentation models are often defined on a closed vocabulary. 
When new classes need to be segmented, it typically requires redefining the vocabulary set, annotating data, and retraining the model.
However, the cost of annotating image segmentation on a large vocabulary is high.
Large-scale image-text pre-training models, such as CLIP~\cite{radford2021clip} and ALIGN~\cite{jia2021align}, have demonstrated strong zero-shot recognition capabilities, sparking interest in transferring these abilities to image segmentation tasks. 
Several studies~\cite{ding2023maskclip,jiao2023maft,yu2024fcclip,jiao2024maftplus} have explored how to adapt mask generators~\cite{cheng2021maskformer,cheng2022mask2former} and pre-trained CLIP to each other to achieve open-vocabulary segmentation at both the semantic and instance levels.
However, our analysis based on two key observations reveals critical shortcomings in existing research that have been consistently overlooked.

% （分）
First, we establish a baseline method enabling CLIP to perceive mask categories by referencing prior methodologies~\cite{xu2023san,yu2024fcclip}. 
To evaluate the impact of mask quality on inference performance, we conduct comparative experiments using generated versus ground-truth masks across various pre-trained CLIP models. 
As shown in \cref{subfig:oracle_mask_results}, generated masks exhibit inferior performance compared to ground-truth masks due to their low segmentation accuracy, semantic ambiguity, and excessive redundancy. 
Our experiments further demonstrate that employing higher-quality masks during training consistently yields superior performance under identical inference protocols.

Second, to validate the strong generalization capability of existing mask generators, 
we evaluate multiple architectures pre-trained on the COCO dataset~\cite{lin2014mscoco} and evaluate on the ADE20K dataset~\cite{zhou2017ade20k} with a different vocabulary set. 
Through IoU-based bipartite matching, we reassign ``ground-truth categories'' to predicted masks to determine whether these generators produce masks for unseen categories. 
As illustrated in \cref{subfig:oracle_cls_results}, the performance could be substantially improved if mask classification were perfect. 
This discrepancy suggests that existing mask generators inherently produce numerous valid masks for unseen categories, but their potential is severely limited by erroneous mask classification.

% （总）基于预实验，我们将在训练中去除对掩码生成器的依赖，并试图直接用高质量的图像分割数据微调CLIP。
Based on these observations, we propose to eliminate the reliance on mask generators during training 
and directly fine-tune CLIP using high-quality image segmentation data.
However, since image segmentation data is typically defined on closed vocabularies, mitigating the issue of CLIP overfitting the training data after fine-tuning poses an important challenge.
Existing approaches either integrate a frozen CLIP during inference~\cite{xu2022zsseg,xu2023odise,yu2024fcclip}
or employ a frozen teacher model for distillation~\cite{jiao2023maft,jiao2024maftplus}.
The former may limit the model's performance improvement, 
while the latter often requires careful tuning of distillation strategies and hyperparameters.
To address this issue more effectively, we analyze the causes of modality misalignment in prior works 
and propose that the introduction of additional parameters should adhere to the consistent alignment principle, 
thereby preserving CLIP's original embedding space throughout the alignment optimization process.
The outcome of our explorations is a new CLIP fine-tuning framework, coined as \Ours.

Experimental results demonstrate that \Ours{} can significantly enhance CLIP's masked classification capability across datasets spanning diverse domains.
Additionally, \Ours{} can be used with different mask generators to achieve open-vocabulary segmentation at the semantic- or instance-level.
Compared to recent state-of-the-art methods, 
\Ours{} offers lower training cost in terms of training time and memory usage, while achieving even better performance.
We hope that the universality and efficiency of \Ours{} could inspire further research in open-vocabulary segmentation.

%% file: sec_arxiv/2_related.tex
\section{Related Work}

\myPara{Mask-based segmentation.}
Mask-based segmentation is a technique that views image segmentation as the process of mask generation and mask classification, which is also known as region-based segmentation.
To address instance and panoptic segmentation,~\cite{he2017maskrcnn,wang2020solo,wang2021max,zhang2021knet}
proposed various mask generators.
Recently, \cite{cheng2021maskformer} unified the
semantic, panoptic and instance segmentation into a single architecture.
MaskFormer~\cite{cheng2021maskformer} uses learnable queries to interact with image features, which are then decoded as masks and class scores.
Mask2Former~\cite{cheng2022mask2former} further accelerates the convergence of MaskFormer and reduces memory usage through 
multi-scale decoder design, using masks from shallow decoders as priors for deeper layers, and employing sampling-based optimization.
Current mask-based OVS methods need almost to retrain an open-vocabulary mask generator based on Mask2Former~\cite{xu2023masqclip,xu2023odise,yu2024fcclip}.
In contrast, our method can directly use the Mask2Former trained on a closed vocabulary during inference, 
reducing training cost and increasing flexibility in usage.

\begin{figure*}[t]
  \centering
  \begin{subfigure}[c]{0.32\textwidth}
      \centering
      \includegraphics[width=\textwidth]{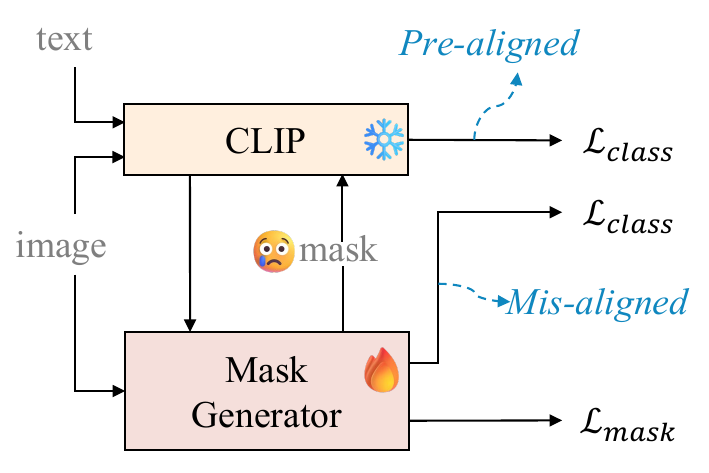}
      \caption{Adapt Mask Generator to CLIP}
      \label{subfig:original_segmenter_adaptation}
  \end{subfigure}
  \hfill
  \begin{subfigure}[c]{0.32\textwidth}
      \centering
      \includegraphics[width=\textwidth]{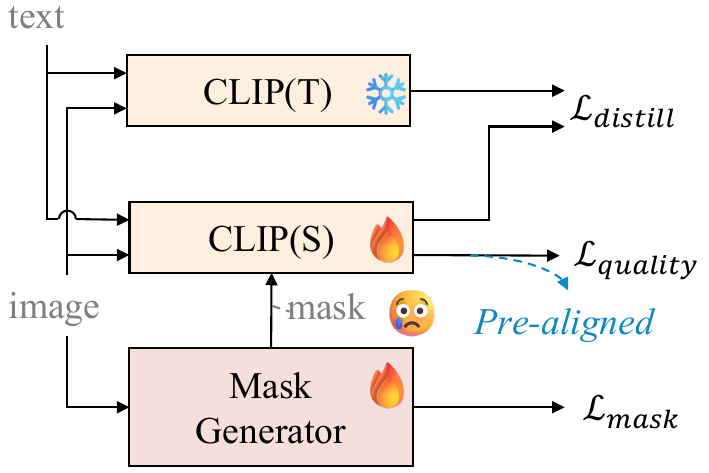}
      \caption{Adapt CLIP to Mask Generator}
      \label{subfig:original_clip_adaptation}
  \end{subfigure}
  \hfill
  \begin{subfigure}[c]{0.32\textwidth}
    \centering
    \includegraphics[width=\textwidth]{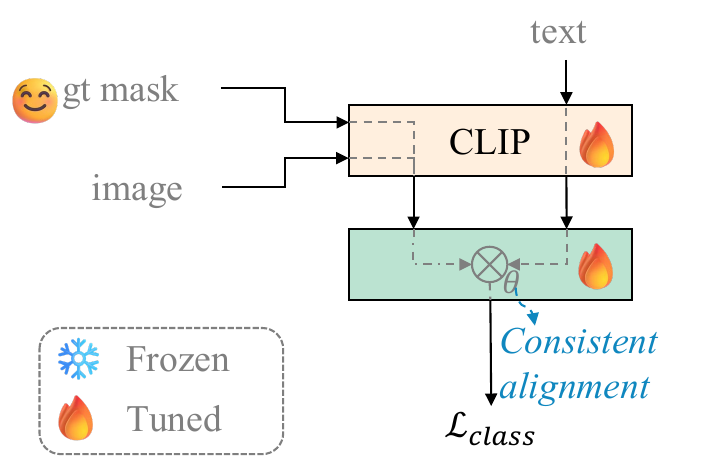}
    \caption{\Ours}
    \label{subfig:our_clip_adaptation}
  \end{subfigure}
  % \vspace{-1.2em}
  \caption{
    Comparison of training pipeline between previous mask-based OVS methods.
    $\mathcal{L}$ denotes the loss function.
    \protect{\subref{subfig:original_segmenter_adaptation}} adapts the mask generator to CLIP and avoids overfitting by freezing CLIP.
    \protect{\subref{subfig:original_clip_adaptation}} adapts CLIP to the mask generator and avoids overfitting by distillation. ``T'' and ``S'' denote the teacher and student models, respectively.
    Our method \protect{\subref{subfig:our_clip_adaptation}} abandons the mask generator and avoids overfitting by consistency alignment.
  }
  \label{fig:training_components}
\end{figure*}

% OVS
\myPara{Open-vocabulary segmentation.}
Depending on the data used, existing methods have developed into different paradigms.
The first paradigm offers many heuristic solutions based on observations of CLIP without additional training~\cite{zhou2022extract,wang2023sclip,li2023clipsurgery,sun2024clipasrnn,lan2024clearclip,lan2024proxyclip}.
The second paradigm utilizes image-text pair data~\cite{thomee2016yfcc100m,changpinyo2021cc12m} to train OVS models under \textit{weak supervision}~\cite{xu2022groupvit,luo2023segclip,xu2023ovsegmentor,zhang2023pgseg,mukhoti2023pacl,cha2023tcl,liu2024mgca}.
Due to the lack of precise segmentation locations, these methods suffer from poor segmentation quality.
The third paradigm enhances segmentation quality by incorporating image-mask data~\cite{kirillov2023sam} and knowledge from SAM~\cite{wang2024uniovseg,yuan2024ovsam}.
However, the use of SAM not only increases inference cost but also relies on users or other detectors to provide high-quality prompts.
The fourth paradigm seeks to achieve mutual adaptation between the segmentation model and CLIP through 
\textit{fully supervised} training on image segmentation datasets containing image, mask, and category triplets, 
thereby producing an open-vocabulary segmentation model.
Based on the type of the segmentation model,
these methods can be categorized into mask-based approaches~\cite{ding2022zegformer,xu2022zsseg,ghiasi2022openseg,han2023deop,liang2023ovseg,qin2023freeseg,jiao2023maft,ding2023maskclip,xu2023masqclip,xu2023san,xu2023odise,yu2024fcclip,liu2024scan,jiao2024maftplus}
and pixel/patch-based ones~\cite{li2022lseg,zhou2023zegclip,li2024cascade,cho2024catseg,xie2024sed}. 
Pixel-based OVS methods typically focus on semantic segmentation and cannot perform instance-level segmentation independently.
Additionally, some unique approaches to achieving OVS involve 
image retrieval~\cite{shin2022reco,barsellotti2024fossil}, feature distillation~\cite{chen2023zeroseg,wu2024clipself}
or model merging~\cite{wang2024samclip}.
Some unified large models also incorporate the functionality of OVS~\cite{zou2023xdecoder,shen2024ape}.
This paper focuses on efficiently transferring CLIP on limited segmentation datasets 
to improve its performance on semantic- and instance-level open-vocabulary segmentation tasks.

%% file: sec_arxiv/3_method.tex
\section{Method}

\subsection{Revisiting Image-Text Embeddings Alignment for OVS}\label{sec:revisit}
We first revisit the roles of the visual encoder (CLIP-V) and text encoder (CLIP-T) in the pretrained CLIP model. 
Denote the image embeddings from CLIP-V as $E_i \in \mathbb{R}^{1 \times D}$, where $D$ represents the dimension of embedding space.
To create text embeddings, a vocabulary set of $K$ categories is transformed into the corresponding sentences using templates like ``A photo of \{\}'', and then encoded by CLIP-T, yielding $E_t \in \mathbb{R}^{K \times D}$. 
The original CLIP model maps both images and text into an aligned embedding space. 
In this space, image-text pairs maintain semantic alignment, which we describe as ``pre-alignment'' and can often be measured using cosine similarity $\langle E_i, E_t \rangle$.
In open-vocabulary segmentation, preserving the pre-alignment property is crucial for adapting CLIP to zero-shot mask classification, as it enhances the alignment of mask and text embeddings.

Previous mask-based OVS methods obtain pre-aligned mask embeddings and text embeddings in two ways.
As illustrated in \cref{subfig:original_segmenter_adaptation}, 
the first way aims to train a mask generator that adapts to CLIP,
with the generalization mainly derived from the original frozen parameters of CLIP~\cite{xu2023san,ding2023maskclip}.
However, attempts to align the embedding spaces of the mask generator and CLIP often fail~\cite{xu2022zsseg,yu2024fcclip}.
As illustrated in \cref{subfig:original_clip_adaptation}, 
the second way make CLIP sensitive to the quality of  generated masks during training to enhance performance 
and avoids overfitting through distillation~\cite{jiao2023maft,jiao2024maftplus}.

We identify two limitations of these two types of methods. 
First, they rely on noisy generated masks, 
which can degrade the quality of the region representations (as shown by the vertical comparisons in \cref{subfig:oracle_cls_results}).
This degradation can weaken the alignment between region-specific image and text embeddings.
% thereby weakening the semantic alignment between the image and text embeddings.
%
Second, there is a lack of an intuitive explanation and a straightforward solution for learning better alignment without severe overfitting. 
Moreover, freezing CLIP limits its transferability from global to local features, while distillation typically requires carefully designed strategies and incurs additional training cost.

To overcome these limitations, we propose a new training pipeline for OVS, named \Ours,
which, to our knowledge, is the first method that attempts to leverage ground truth (GT) masks to obtain region features instead of relying on the mask generator during training.  
Furthermore, we propose the principle named ``consistency alignment'' to alleviate overfitting when adding extra parameters to optimize alignment.
We will give detailed descriptions in what follows.

\begin{figure*}[t]
  \centering
  \setlength{\abovecaptionskip}{2pt}
  \includegraphics[width=\textwidth]{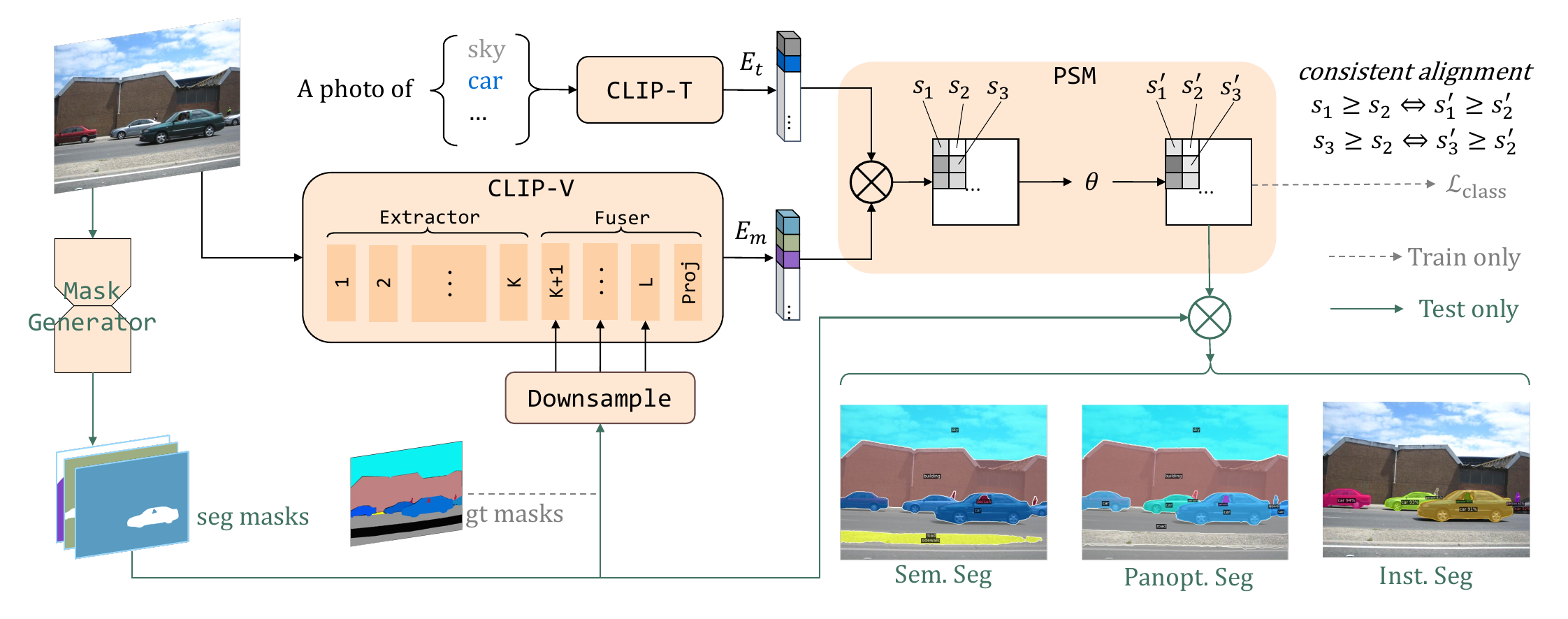}
  \caption{
    Detailed framework of \Ours{} for OVS tasks.
    The PSM represents the parameterized similarity modeling, which is designed under the principle of consistency alignment.
    The mask generator is only used during inference, and can be flexibly replaced.
    }
\label{fig:framework}
\end{figure*}

\subsection{A New Fine-tuning Framework for CLIP}\label{sec:framework}
In this subsection, we detail how to use our framework to fine-tune CLIP 
and integrate it with an off-the-shelf mask generator. The overall framework is illustrated in \cref{fig:framework}.

\myPara{Extracting mask embeddings.}
Taking CLIP-V with an $L$-layer Transformers as an example, inspired by SAN~\cite{xu2023san},
we posit that the layers from $1$ to $K$ predominantly perform feature extraction (denoted as the \textbf{Extractor}), 
while layers from $K+1$ to $L$ primarily conduct spatial fusion (referred to as the \textbf{Fuser}).

% The mask is inserted after the last feature $F_n \in\mathbb{R}^{C\times H\times W}$ of the encoder is obtained, 
% converting the original global aggregation into an aggregation within the mask's effective region as follows:
% 对于输入融合器的第 $l$ 层的 N 个图像tokens $F^{(l)} \in \mathbb{R}^{N\times C}$，我们将使用 $Q$ 个展平的图像掩膜 $M\in[0,1]^{Q\times N}$ 作为条件，以将融合范围从全局聚合转变为局部聚合。
For the $N$ image tokens $F^{(l)} \in \mathbb{R}^{N\times C}$ that input to the $l$-th layer of the fusion module, 
we introduce $Q$ flattened image masks $M\in[0,1]^{Q\times N}$ as conditions to transform the fusion scope from global to local.
% 具体来说，如式(1)所示，我们基于掩码计算一组权重 $\phi^l$，对 $F^{(l)}$ 进行加权求和，再通过该层的通道映射函数 $\operatorname{Proj}&{(l)}$ 得到属于该掩码的token $E_m^{(l+1)}\in \mathbb{R}^{Q\times C}$.
As formally delineated in \cref{eq:mask_embedding}, we compute mask-conditioned modulation weights $\phi^l$ to perform weighted aggregation on $F^{(l)}$, followed by the layer-wise channel projection operator $\mathrm{Proj}^{(l)}$, thereby generating mask-specific tokens $E_m^{(l+1)}\in \mathbb{R}^{Q\times C}$.
\begin{equation}\label{eq:mask_embedding}
  E_m^{(l+1)} = \operatorname{Proj}^{(l)}\left(\phi^{(l)}(M) \cdot F^{(l)}\right),~\text{where}~K < l < L.
\end{equation}

By leveraging the input projections $q$ and $k$ from the original attention module, we incorporate the mask into the weighting function through \cref{eq:phi}, thereby constraining the weight distribution within the mask's active regions.
\begin{equation}\label{eq:phi}
  \phi^{(l)}(M) = \operatorname{softmax}\left ( \frac{q(E_m^{(l)})k(F^{(l)})^T}{\sqrt{d}} + \alpha M_\text{th} \right ),
\end{equation}
where $\sqrt{d}$ is the attention scaling factor, $\alpha \in (0, +\infty)$ is an learnable parameter, $M_\text{th}$ is the result of setting all values in $M$ that are less than $\max(M)/2$ to $-\infty$.

When initializing $E_m^{(K+1)}$ with the CLS token in layer $K+1$, $E_m^{(L)}$ and $E_t$ exhibit pre-aligned properties. 
The first two rows of \cref{tab:maskacc} quantitatively demonstrate the mask classification performance of pre-trained CLIP under this initialization paradigm.

\begin{figure}[t]
  \centering
  \setlength{\abovecaptionskip}{2pt}
  \includegraphics[width=\linewidth]{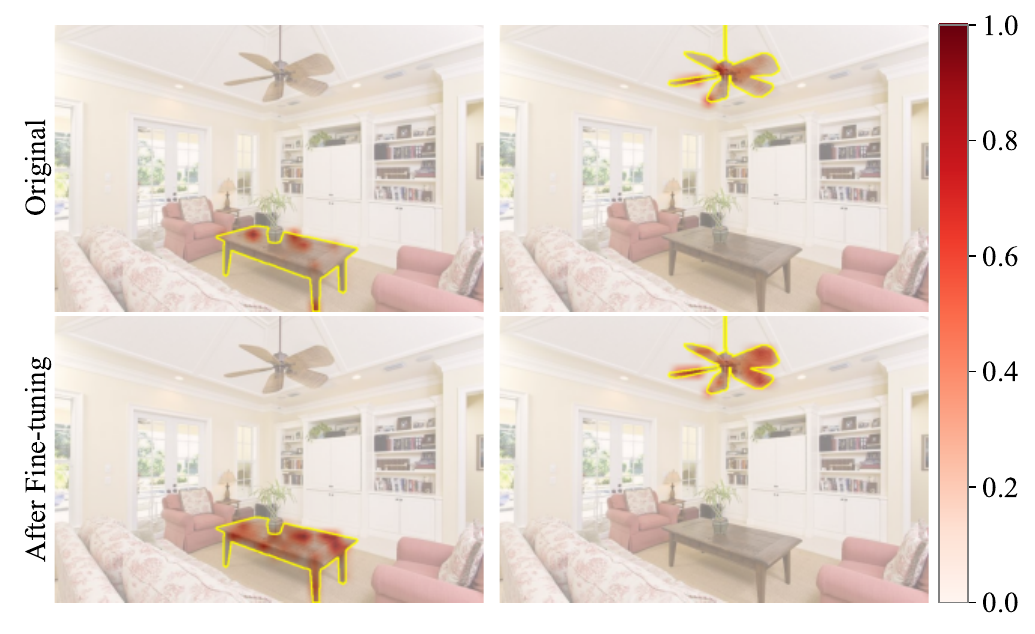}
  \caption{
    Visualization of the weight function. 
    The regions of interest are enclosed by yellow contours, and the deeper the red color at a position, the higher its importance to the region.
  }
  \label{fig:vis_attn}
\end{figure}

\cref{fig:vis_attn} visualizes the normalized weight function $\phi$, which reveals the contributions of image tokens at different spatial positions to the generation of the current mask representation. 
After fine-tuning, more image tokens are incorporated into the construction of their corresponding mask representations.

% insert figures/vis_attn

\myPara{Parameterized similarity modeling with consistency alignment.}
Given the limited capacity of the original CLIP model in characterizing localized regions, 
a straightforward approach involves fine-tuning the original CLIP parameters alongside additional learnable parameters to reformulate the similarity metric between $E_m$ and $E_t$. 
We introduce a class of methods termed Parameterized similarity modeling (PSM) for this purpose.

However, 
when incorporating the mask generator into similarity modeling, 
the mask embeddings produced by the mask generator align almost exclusively with CLIP's text embeddings on the training set, 
resulting in the misalignment illustrated in \cref{subfig:original_segmenter_adaptation}.
 We first analyze the cause of this overfitting to develop our similarity modeling approach.

To facilitate discussion, we simplify this approach as follows.
Let $t \in \mathbb{R}^D$ denote a normalized text embedding from CLIP, and let $m_1, m_2 \in \mathbb{R}^D$ denote different mask embeddings.
The initial similarity is defined as $s = m^T t$.
Now, before computing the inner product across modalities, the mask embeddings are updated using a parameter $\Theta \in \mathbb{R}^{D \times D}$,
resulting in a new similarity $r = (\Theta m)^Tt$.
Although $r$ can be optimized to be close to the target, an unintuitive phenomenon may arise:
After mask embeddings are updated via gradient descent,
the order of $r_1$ and $r_2$ is closer to the target, but
the order of $s_1$ and $s_2$ may deviate further from the target. 
This phenomenon is illustrated by a toy example in \cref{sec:phenomenon} in supplementary material.

Consequently, the original similarity relationships between CLIP's embeddings are disrupted. 
Our experiments demonstrate that the additional parameter $\Theta$ fails to achieve CLIP's level of generalization when fine-tuning on limited image segmentation data, resulting in overfitting.
Building upon the aforementioned analysis, we identify inconsistent alignment phenomena in PSM that prone to overfitting. 
This implies that preserving CLIP's original similarity ordering to the greatest extent possible during optimization enables more effective utilization of additional parameters for alignment enhancement. 
We formalize this critical finding as the \textit{consistency alignment} principle in PSM.

Specifically, for all mask embeddings $E_m\in \mathbb{R}^{Q\times C}$ extracted from an image 
and the text embeddings $E_t \in \mathbb{R}^{K\times C}$ of $K$ dataset categories, their similarity matrix 
$S = E_m E_t^T$ was projected through an over-parameterized linear projection along a newly created orthogonal dimension relative to both $Q$ and $K$ dimensions. 
Finally, cross-entropy loss was computed based on category annotations of each mask, guided by the optimized similarity map.

\myPara{Integrating with off-the-shelf mask generators.}
As the training process is entirely decoupled from the mask generator, 
we can flexibly employ various types of mask generators during inference to achieve semantic- or instance-level open-vocabulary segmentation.

Compared to previous OVS methods, \Ours{} offers the following several advantages.
First, we use ground-truth masks to directly fine-tune CLIP during training, which are less noisy,
thus improving alignment quality and reducing training cost.
Second, from the perspective of avoiding overfitting, 
the consistency alignment principle provides more optimization space than freezing CLIP~\citep{xu2023san,yu2024fcclip},
and does not rely on carefully designed distillation strategies~\citep{wu2024clipself,jiao2024maftplus},
resulting in faster convergence with a single optimization objective.

%% file: sec_arxiv/4_experiments.tex
\section{Experiments}
\subsection{Experimental Setup}
\myPara{Baseline.} 
In this section, we employ pre-trained models from EVA02 CLIP~\cite{sun2023evaclip} with ViT architecture. 
% Experimental results for CLIP variants with alternative architectures are provided in Appendix \todo.  
Following the framework outlined in \cref{sec:framework}, we configure $L-K = 2$ with $\alpha$ initialized to $e^{-5}$.  
The second row of \cref{tab:maskacc} demonstrates the mask classification accuracy of this baseline implementation.

For open-vocabulary image segmentation evaluation, we adopt FC-CLIP (corresponding to \cref{subfig:original_segmenter_adaptation}) and MAFT+ (corresponding to \cref{subfig:original_clip_adaptation}) as baseline approaches.  
Their baseline performance metrics are documented in \cref{tab:ovss} and \cref{tab:ovps}.  
By retaining the mask generators from both frameworks while replacing the mask classification component with \Ours, we demonstrate the performance enhancement in overall OVS capabilities through improved mask classification accuracy.

\myPara{Datasets and Metrics.}
Aligning with established OVS methodologies, we fine-tune our model on the COCO Stuff dataset~\cite{caesar2018cocostuff} with its fixed 171-category vocabulary
and evaluate on datasets with different vocabularies:  ADE20K (150-class and 847-class taxonomies)~\cite{zhou2017ade20k}, Pascal Context (459-class and 59-class taxonomies)~\cite{mottaghi2014context}, and PASCAL VOC (20 foreground categories excluding background)~\cite{everingham2010pascal}.  
The mean Intersection over Union (mIoU) metric serves as our primary evaluation criterion for OVS performance. 
Furthermore, we quantitatively demonstrate performance enhancements in both panoptic and instance segmentation tasks on ADE20K through established metrics: panoptic quality (PQ), segmentation quality (SQ), recognition quality (RQ) and mean average precision (mAP).

To demonstrate the generalizable improvement of \Ours{} in mask classification in various scenarios, we perform evaluations on datasets from the MESS benchmark~\cite{MESSBenchmark2023}, where the accuracy of the top-1 ground truth mask (maskAcc) serves as the primary metric to quantify the capability of zero shot mask classification.

% Further details on the datasets are provided in the Appendix \todo.

\myPara{Implementation Details.}
Our implementation is built on the Detectron2 framework~\cite{wu2019detectron2}.
During training, the number of image tokens $N=32^2$. The AdamW optimizer is employed with a batch size of 4 for 20K iterations. 
The initial learning rate is set to $10^{-4}$ and decayed using cosine annealing.
Following the setting of CAT-Seg~\cite{cho2024catseg}, we only train the linear projection parameters of the query and value in the CLIP attention modules, and set their learning rate $10^2$ times the overall learning rate.
The over-parameterized dimension of the linear layer in PSM is set to 768.

% During inference, images are resized by scaling the shorter edge to 392 pixels while preserving the original aspect ratio for processing by CLIP-V, 
% while the images input to the mask generator maintain their original configuration of being resized with the shorter edge scaled to 800 pixels.
During inference, while maintaining the original aspect ratio, images are resized with shorter edges scaled to 392 and 800 pixels respectively for CLIP-V and Mask Generator inputs.
Similar to previous works~\cite{xu2023odise,yu2024fcclip}, we also ensemble the class scores of the mask generator and \Ours{} on the training categories of the mask generator.

\begin{table*}[tbp]
    \centering
    \setlength{\abovecaptionskip}{2pt}
    \resizebox{0.9\textwidth}{!}{%
    \begin{tabular}{lccccccccccc}
    \toprule
                                        & \textbf{A-847} & \textbf{PC-459} & \textbf{A-150} & \textbf{PC-59} & \textbf{Stuff} & \textbf{Citys} & \textbf{General} & \textbf{Earth} & \textbf{Medical} & \textbf{Engineer} & \textbf{Agriculture} \\
    \midrule
    \emph{\textbf{Pretrained}} \\
    OpenAI CLIP~\cite{radford2021clip}  & 32.0    & 44.8   & 51.0    & 57.9  & 46.2       & 46.5  & 54.2    & 61.1  & 61.0    & 36.7     & 51.0        \\
    EVA02 CLIP~\cite{sun2023evaclip}    & 35.2    & 44.8   & 52.7    & 54.6  & 45.0       & 44.9  & 56.9    & 60.5  & 61.7    & 33.8     & 52.4        \\
    \emph{\textbf{Finetuned on COCO}} \\
    CLIPSelf~\cite{wu2024clipself}      & 33.6    & 49.0   & 56.1    & 67.5  & 50.6       & 52.1  & 52.5    & 45.7  & 55.5    & 45.3     & 56.3        \\
    CAT-Seg~\cite{cho2024catseg}        & 33.5    & 50.7   & 62.3    & 80.6  & 63.0       & 65.3  & 60.7    & 66.9  & 62.0    & 34.9     & 64.4        \\
    \rowcolor{GrayBack}
    \Ours{}                             & \textbf{38.4}    & \textbf{56.4}   & \textbf{67.0}    & \textbf{85.2}  & \textbf{67.8}       & \textbf{71.0}  & \textbf{67.9}    & \textbf{68.6}  & \textbf{74.7}    & \textbf{50.3}     & \textbf{65.5}        \\
    \bottomrule
    \end{tabular}%
    }
    \caption{
        Comparison of the classification accuracy of ground-truth masks (maskAcc) with different methods on multiple datasets. 
        The same architecture (ViT-L/14) and input resolution (short side length 392) are used for all methods.
        ``General'', ``Earth'', ``Medical'', ``Engineer'', and ``Agriculture'' represent mean mask accuracy across five domain datasets in the MESS benchmark~\cite{MESSBenchmark2023}, with detailed scores for each individual dataset presented in the supplementary material.
        }
    \label{tab:maskacc}
\end{table*}

\begin{table*}[tbp]
    \centering
    \setlength{\abovecaptionskip}{2pt}
    \resizebox{0.9\linewidth}{!}{%
    \begin{tabular}{llllllll}
    \toprule
    \textbf{Method}                  & \textbf{CLIP arch.} & \textbf{Training dataset}  & \textbf{A-847} & \textbf{PC-459} & \textbf{A-150} & \textbf{PC-59} & \textbf{PAS-20} \\ 
    \midrule
    \emph{\textbf{Non-mask-based OVS method}}                                                               \\
    OVSeg~\citep{liang2023ovseg}     & ViT-L/14       & COCO Stuff                 & 9.0   & 12.4   & 29.6  & 55.7  & 94.5   \\
    CLIPSelf~\citep{wu2024clipself}  & ViT-L/14       & COCO Stuff                 & 12.4  &   -    & 34.5  & 62.3  &  -     \\
    SED~\citep{xie2024sed}           & ConvNeXt-L     & COCO Stuff                 & 13.7  & 22.1   & 35.2  & 60.6  & 96.1   \\
    CAT-Seg~\citep{cho2024catseg}    & ViT-L/14       & COCO Stuff                 & \uline{16.0}  & \uline{23.8}   & \uline{37.9}  & \textbf{63.3}  & \uline{97.0}   \\ 
    \midrule
    \emph{\textbf{Mask-based OVS method}}                                                                \\
    MaskCLIP~\citep{ding2023maskclip}& ViT-L/14       & COCO Panoptic              & 8.2   & 10.0   & 23.7  & 45.9  & -      \\
    MasQCLIP~\citep{xu2023masqclip}  & ViT-L/14       & COCO Panoptic              & 10.7  & 18.2   & 30.4  & 57.8  & -      \\
    ODISE~\citep{xu2023odise}        & ViT-L/14       & COCO Panoptic              & 11.1  & 14.5   & 29.9  & 57.3  & -      \\
    SAN~\citep{xu2023san}            & ViT-L/14       & COCO Stuff                 & 12.4  & 15.7   & 32.1  & 57.7  & 94.6   \\
    SCAN~\citep{liu2024scan}         & ViT-L/14       & COCO Stuff                 & 14.0  & 16.7   & 33.5  & 59.3  & \textbf{97.2}   \\
    FC-CLIP~\citep{yu2024fcclip}     & ConvNeXt-L     & COCO Panoptic              & 14.8  & 18.2   & 34.1  & 58.4  & 95.4   \\
    MAFT+~\citep{jiao2024maftplus}   & ConvNeXt-L     & COCO Stuff                 & 15.1  & 21.6   & 36.1  & 59.4  & 96.5   \\
    \rowcolor{GrayBack}
    \Ours{} w/ FC-CLIP               & ViT-L/14       & COCO Stuff                 & 15.4\redsub{+0.6}  & 21.3\redsub{+3.1}   & 37.1\redsub{+3.0}  & 62.6\redsub{+4.2}  & 96.4\redsub{+1.0}   \\
    \rowcolor{GrayBack}
    \Ours{} w/ MAFT+                 & ViT-L/14       & COCO Stuff                 & \textbf{16.8}\redsub{+1.7}  & \textbf{23.9}\redsub{+2.3}   & \textbf{38.2}\redsub{+2.1}  & \uline{62.5}\redsub{+3.1}  & 96.8\redsub{+0.3}     \\
    \bottomrule
  \end{tabular}
  }
  \caption{
      Comparisons with previous methods on open-vocabulary semantic segmentation task. 
    }
  \label{tab:ovss}  
\end{table*}

\subsection{Main results}

\myPara{Mask classification.}
\cref{tab:maskacc} presents the mask classification accuracy of CLIP models from different sources when processing input images and ground-truth masks at identical resolutions. 
In addition to the two pre-trained CLIP variants, we also demonstrate the performance of CLIP fine-tuned on COCO through distinct approaches:
CLIPSelf~\cite{wu2024clipself} learned via patch-level feature distillation and CAT-Seg~\cite{cho2024catseg} trained with pixel-level cost aggregation. 
During inference, CLIPSelf performs mask pooling on the dense features of the final layer. 
Given that CAT-Seg operates as a pixel-level segmentation model, we leverage its CLIP parameters while maintaining consistency with the baseline methods by employing identical mask classification methodology.
By directly utilizing higher-quality masks to fine-tune CLIP while adhering to consistent alignment principles, 
\Ours{} achieves significant improvements over baselines across multi-domain datasets, outperforming other fine-tuning approaches.

\myPara{Open-vocabulary semantic segmentation.}
\cref{tab:ovss} demonstrates the mIoU of \Ours{} when employing mask generators from FC-CLIP~\cite{yu2024fcclip} or MAFT+~\cite{jiao2023maft}. 
Compared to MAFT+, our approach achieves mIoU improvements of +1.7, +2.3, +2.1, +3.2, and +0.3 points on the A-847, PC-459, A-150, PC-59, and PAS-20 datasets, respectively. 
These results indicate that our method more effectively harnesses the potential of existing mask generators through better mask classification capabilities.

\myPara{Open-vocabulary panoptic and instance segmentation.}
\cref{tab:ovps} demonstrates that our method can be effectively integrated with existing instance-level mask generators. 
Furthermore, the enhanced classification capabilities of \Ours{} also improve the performance across these segmentation paradigms.

\begin{table}[t]
  % \resizebox{\linewidth}{!}{%
  \centering
  \setlength{\tabcolsep}{7pt}
  \setlength{\abovecaptionskip}{2pt}
  \small
  \begin{tabular}{lcccc}
    \toprule
    % \textbf{Method}                        & \textbf{PQ} & \textbf{AP} \\ 
    % \multirow{2}{*}{\textbf{Method}}       & \multicolumn{2}{c}{\textbf{A-150}} & \multicolumn{2}{c}{\textbf{Citys}} \\
    % \cmidrule(lr){2-3} \cmidrule(lr){4-5}
    \textbf{Method}                        & \textbf{PQ} & \textbf{SQ} & \textbf{RQ} & \textbf{AP} \\

    \midrule
    MaskCLIP~\citep{ding2023maskclip}      & 15.1        & 70.5        & 19.2        & 6.2         \\
    MasQCLIP~\citep{xu2023masqclip}        & 23.3        &   -         &   -         &  -          \\
    ODISE~\citep{xu2023odise}              & 22.6        &   -         &   -         & 14.4        \\
    FC-CLIP~\citep{yu2024fcclip}           & 26.8        & 71.5        & 32.2        & 16.8        \\
    MAFT+~\citep{jiao2024maftplus}         & 27.1        & 73.5        & 32.9        &  -          \\
    \rowcolor{GrayBack}
    \Ours{} w/ FC-CLIP                     & 27.7        & 72.0        & 33.6        & \textbf{17.3}        \\
    \rowcolor{GrayBack}
    \Ours{} w/ MAFT+                       & \textbf{28.1}        & \textbf{74.0}        & \textbf{34.7}        & -        \\
    \bottomrule
    \end{tabular}%
  % }
  \caption{
      Comparisons with previous methods on open-vocabulary panoptic and instance segmentation on A-150. 
    }
  \label{tab:ovps}
\end{table}

\subsection{Ablation Studies} 
\label{sec:ablation}
The ablation studies in \cref{tab:psm} and \cref{tab:prior_quality} are conducted using ViT-B/16 models fine-tuned on COCO Panoptic~\cite{lin2014mscoco}, 
whereas those presented in \cref{tab:generators} and \cref{tab:data_size} employed ViT-L/14 models fine-tuned on COCO Stuff~\cite{caesar2018cocostuff}.

\myPara{Importance of consistency alignment.}
We use different parameterized similarity modeling methods to illustrate the importance of the consistent alignment principle.
\cref{tab:psm} elucidates the relationship between consistency alignment and overfitting through PSM application analysis during inference. 
We evaluate alignment consistency by toggling PSM usage after fine-tuning, 
while partitioning validation set categories into seen (S) and unseen (U) subsets to compute mIoU(S) and mIoU(U) respectively. 
Severe overfitting is identified when optimized mIoU(U) significantly underperforms its no-fine-tuning baseline. 
The table substantiates that the $\operatorname{Linear}\langle E_m, E_t \rangle$ approach preserving consistency alignment achieves simultaneous improvements in both mIoU(S) and mIoU(U), 
demonstrating effective generalization maintenance.

\begin{table}[t]
  \centering
  \small
  % \renewcommand{\arraystretch}{0.98}
  % \resizebox{\linewidth}{!}{%
  \setlength{\abovecaptionskip}{2pt}
    \begin{tabular}{ccccc}
    \toprule
    \textbf{PSM}                                                       & \textbf{Use}      & \textbf{mIoU(S)} & \textbf{mIoU(U)} & \textbf{mIoU} \\
    \midrule
    No fine-tune                                                       & -                                           & 22.6         & 12.8         & 17.0      \\ \midrule
    \textcolor{RedFore}{\multirow{2}{*}{$\langle \operatorname{Linear}(E_m), E_t \rangle$}} & \xmark                 & 36.6         & 21.7         & 28.0      \\
                                                                      & \cmark                                       & 40.4         & 6.2          & 20.8      \\ \midrule
    \textcolor{RedFore}{\multirow{2}{*}{$\langle E_m, \operatorname{Linear}(E_t) \rangle$}}  & \xmark                & 36.1         & 21.4         & 27.7      \\
                                                                      & \cmark                                       & 41.2         & 6.6          & 21.4      \\ \midrule
    \textcolor{BlueFore}{\multirow{2}{*}{$\operatorname{Linear}\langle E_m, E_t \rangle$}}   & \xmark                & 44.8         & 25.1         & 33.5      \\
                                                                      & \cmark                                       & 45.3         & 25.3         & 33.8      \\
    \bottomrule
    \end{tabular}%
    % }
    \caption{
        The phenomenon of (in)consistent alignment when fine-tuning with different PSMs.
        The PSMs prone to overfitting (\textcolor{RedFore}{red}) paradoxically shows better performance when omitting the PSM during inference, revealing the occurrence of inconsistent alignment.
        In contrast, the \textcolor{BlueFore}{blue} PSM maintains consistent alignment while achieving comparable performance both with and without the PSM application, thus effectively mitigating overfitting risks.
    }
    \label{tab:psm}
\end{table}

\myPara{Impact of prior quality during training.}
We refer to the regional information fed into Fuser of CLIP-V as ``priors'' 
and investigate how the quality of these priors affects fine-tuning performance. 
We experiment with pixels, generated masks that match GT masks according to IoU, 
GT masks, and the GT boxes as their minimum bounding rectangles.
As shown in \cref{tab:prior_quality},
using GT masks can achieve the best performance.
Using the generated masks as prior yields lower performance compared to the GT masks, 
indicating that incorporating the mask generator during training actually hinders CLIP fine-tuning. 
Both GT masks and GT boxes priors outperform pixel priors, 
suggesting that fine-tuning CLIP from global recognition to region-level recognition is less challenging than fine-tuning it to pixel-level recognition.

\begin{table}[htbp]
  \centering
  \small
  \setlength{\abovecaptionskip}{2pt}
  \setlength{\tabcolsep}{7pt}
  % \resizebox{\linewidth}{!}{
      \begin{tabular}{lcccc}
      \toprule
      \textbf{Prior Type}  & \textbf{mIoU(S)} & \textbf{mIoU(U)}  & \textbf{mIoU}  & \textbf{maskAcc}\\ 
      \midrule
      Pixels               & 43.2             & 21.4              & 30.6           & 53.3 \\ 
      Gen. masks            & 43.8             & 23.8              & 32.3           & 52.2 \\ 
      GT boxes             & 44.4             & 24.2              & 32.8           & 57.6 \\ 
      GT masks             & \textbf{45.3}    & \textbf{25.3}     & \textbf{33.8}  & \textbf{58.3} \\ 
      \bottomrule
      \end{tabular}%
    % }
    \caption{Impact of prior quality during training.
    ``Gen. masks'' denotes the matched generated masks. 
    ``GT boxes'' denotes the minimum bounding rectangles of GT masks.
    }
    \label{tab:prior_quality}
\end{table}

\myPara{Impact of different mask generators.}
% 除了 /cref{tab:ovss} 中的 FC-CLIP 和 MAFT+，
Except for FC-CLIP and MAFT+ in \cref{tab:ovss}, various types of mask generators can be integrated with \Ours{}, as shown in \cref{tab:generators}.
We use mask generators from Mask2Former~\citep{cheng2022mask2former} and FC-CLIP~\citep{yu2024fcclip}, 
which are typically considered as closed-vocabulary and open-vocabulary mask generators, respectively.
These mask generators are all trained on COCO Panoptic~\citep{lin2014mscoco} and evaluated on ADE20K~\citep{zhou2017ade20k} using mIoU.
As shown in \cref{tab:generators}, the original performance of ``open-vocabulary'' mask generators do do not significantly outperform ``closed-vocabulary'' ones 
because of the ``mis-alignment'' as shown in \cref{subfig:original_segmenter_adaptation}.
And using our \Ours{} for mask classification can significantly improve the segmentation performance of all mask generators.
This suggests that both open-vocabulary and closed-vocabulary mask generators possess an inherent category-independent mask generation capability that has yet to be fully harnessed.

\begin{table}[]
  % \resizebox{\linewidth}{!}{%
  \centering
  \setlength{\abovecaptionskip}{2pt}
  \setlength{\tabcolsep}{7pt}
  \small
  \begin{tabular}{lccc}
  \toprule
  \textbf{Generator} & \textbf{OV} & \textbf{Gen. Only} & \textbf{w/ \Ours{}} \\
  \midrule
  Mask2Former~(T) & \xmark & 15.7 & 35.6 \\
  Mask2Former~(B) & \xmark & 16.8 & 35.5 \\
  FC-CLIP~(B)     & \cmark & 19.1 & 36.8 \\
  FC-CLIP~(L)     & \cmark & 20.8 & 37.1 \\
  \bottomrule
  \end{tabular}%
  % }
  \caption{
    Impact of different mask generators on OVS performance.
    ``OV'' indicates whether the mask generator is open-vocabulary.
    ``Gen. Only'' indicates the mIoU of the mask generator on A-150.
    ``w/ \Ours'' indicates the mIoU of the mask generator when integrated with \Ours{}.
  }
  \label{tab:generators}
\end{table}

% 训练数据量的影响
\myPara{Impact of training data size.}
Since the goal of our fine-tuning is to transfer CLIP's open-vocabulary recognition capabilities to mask regions, 
rather than learning category-specific pixel groupings from segmentation datasets, 
our method can effectively fine-tune using a smaller amount of image segmentation data.
We randomly sampled varying proportions of data from the COCO-Stuff training set for fine-tuning, 
using the same configuration as the model in \cref{tab:ovss}.
As shown in \cref{tab:data_size}, even with only 1\% of the data (approximately 1K images with annotations), 
the fine-tuned model still demonstrates competitive performance.

\begin{table}[htbp]
  \centering
  \setlength{\abovecaptionskip}{2pt}
  \setlength{\tabcolsep}{7pt}
  \small
  % \resizebox{\linewidth}{!}{%
    \begin{tabular}{cccccc}
    \toprule
    \textbf{Data size} & \textbf{A-847} & \textbf{PC-459} & \textbf{A-150} & \textbf{PC-59}  & \textbf{Stuff} \\ % & \textbf{PAS-20}
    \midrule
    100\% & 38.4 & 56.4 & 67.0 & 85.2 & 67.8 \\
    10\%  & 39.4 & 56.2 & 67.0 & 84.0 & 67.4 \\
    1\%   & 38.7 & 55.4 & 66.3 & 83.5 & 65.5 \\
    0.1\% & 36.3 & 54.5 & 61.2 & 75.5 & 55.7 \\
    \bottomrule
    \end{tabular}%
    % }
  \caption{
    % The impact of the amount of data used for fine-tuning.
    Mask accuracy when fine-tuning with different amounts of data.
    ``data size'' indicates the proportion of data used for fine-tuning relative to the original COCO-Stuff training set.
  }
  \label{tab:data_size}
\end{table}

%% file: sec_arxiv/5_discussion.tex
\newcommand{\addFig}[1]{
  \begin{minipage}{\textwidth}
    \centering
    \includegraphics[width=0.195\textwidth]{figures/vis/sem_seg/#1.jpg}
    \hfill
    \includegraphics[width=0.195\textwidth]{figures/vis/sem_seg/#1_catseg.png}
    \hfill
    \includegraphics[width=0.195\textwidth]{figures/vis/sem_seg/#1_maftp.png}
    \hfill
    \includegraphics[width=0.195\textwidth]{figures/vis/sem_seg/#1_ours.png}
    \hfill
    \includegraphics[width=0.195\textwidth]{figures/vis/sem_seg/#1_gt.png}
  \end{minipage}
}

\begin{figure*}[t]
  % TODO
  \small
  \setlength{\abovecaptionskip}{2pt}
  \begin{minipage}{\textwidth}
    % add a line of four titles: Image, GT, CLIP, Ours
    \hspace{0.06\textwidth}
    Image
    % \hfill
    \hspace{0.15\textwidth}
    CAT-Seg
    % \hfill
    \hspace{0.13\textwidth}
    MAFT+
    % \hfill
    \hspace{0.11\textwidth}
    \Ours{}
    \hfill
    GT
    \hspace{0.08\textwidth}
  \end{minipage}
%   \addFig{ADE_val_00000084}
  \addFig{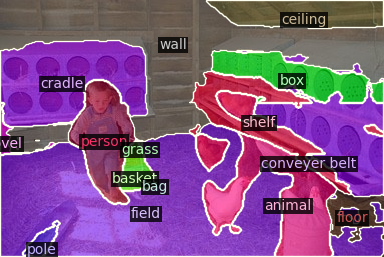}
%   \addFig{ADE_val_00000388}
%   \addFig{ADE_val_00000693}
  \addFig{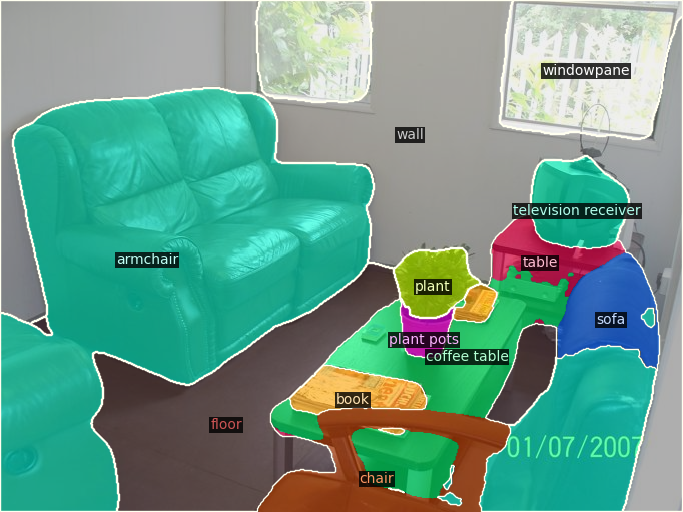}
  \addFig{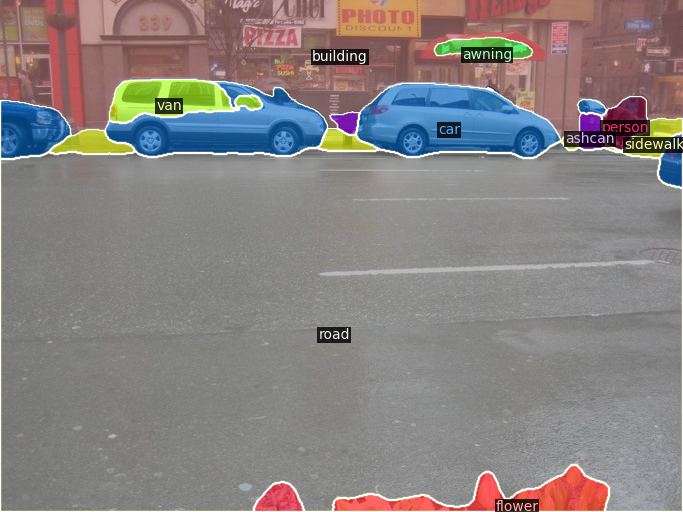}
%   \addFig{ADE_val_00001777}
  \caption{
    Visualizations of open-vocabulary semantic segmentation on ADE20K~\citep{zhou2017ade20k}.
    Our method produces more complete masks than CAT-Seg and fewer biases in mask classification than MAFT+.
  }
  \label{fig:vis_sem_seg}
\end{figure*}

\section{Discussion}
\label{sec:discussion}

\myPara{Qualitative comparison.}
\cref{fig:vis_sem_seg} presents qualitative comparisons of CAT-Seg~\cite{cho2024catseg}, MAFT+~\cite{jiao2024maftplus}, and \Ours{} on open-vocabulary semantic segmentation tasks.
Compared to CAT-Seg, mask-based methods often produce higher-quality masks. 
Compared to MAFT+, our method retains more of the original CLIP's knowledge without generating too many erroneous biases.
Our method also predicts relatively correct semantics for some objects that are not annotated in the GT.

% \newcommand{\insertimage}[2]{
%     \begin{tikzpicture}
%         \node[anchor=south west,inner sep=0] (image) at (0,0) {\includegraphics[width=0.5\textwidth]{#1}};
%         \node[anchor=south east, font=\Large\bfseries, inner sep=3pt] 
%             at (image.south east) {\contour{white}{#2}};
%     \end{tikzpicture}
% }

% \begin{figure}[t]
%     \centering
%     \setlength{\abovecaptionskip}{2pt}
%     \resizebox{\linewidth}{!}{%
%     \begin{tabular}{lr}
%         \insertimage{figures/vis/sem_seg/ADE_val_00000693_clip}{CLIP} & 
%         \insertimage{figures/vis/sem_seg/ADE_val_00000693_maftp}{MAFT+} \\
%         \insertimage{figures/vis/sem_seg/ADE_val_00000693_ours}{\Ours{}} &
%         \insertimage{figures/vis/sem_seg/ADE_val_00000693_gt}{GT}
%     \end{tabular}%
%     }
%     \caption{
%         % 在同样的生成掩码下，使用不同模型（CLIP, MAFT+, MaskCLIP++）进行语义分割的可视化结果 以及 ground truth。
%         Comparative semantic segmentation visualizations using multiple models (CLIP, MAFT+, MaskCLIP++) under identical generated mask conditions, 
%         presented alongside ground truth annotations.
%     }
%     \label{fig:vis_sem_seg_one}
% \end{figure}

\myPara{Segmentation without mask generators.}
Because the goal of our fine-tuning is to enhance the mask classification ability of CLIP,
which indirectly enhances the local representation between different patch representations of the CLIP.
Diverging from \cref{sec:framework}, we apply mask generator-free inference schemes on MaskCLIP++ to demonstrate the enhanced inter-patch relationships, as shown in \cref{tab:inference_wo_mask}.

\begin{table}[tbp]
    \centering
    \setlength{\abovecaptionskip}{2pt}
    \setlength{\tabcolsep}{3pt}
    % \resizebox{\linewidth}{!}{%
    \small
    \begin{tabular}{lccccccc}
    \toprule
    \textbf{Method} & \textbf{Load}                               & \textbf{PC-459} & \textbf{A-150} & \textbf{PC-59} & \textbf{Stuff} & \textbf{Citys} \\ % % & \textbf{A-847} 
    \midrule % & -              
    \multirow{2}{*}{SCLIP~\citep{wang2023sclip}}         & \xmark & 7.7   & 16.1 & 34.2 & 22.4 & 32.2 \\                                                     % & -              
                                                         & \cmark & 9.5  & 19.3 & 40.8 & 29.0 & 35.9 \\ \midrule                                            % & -              
    \multirow{2}{*}{ClearCLIP~\citep{lan2024clearclip}}  & \xmark & 7.9  & 16.7 & 36.0 & 23.9 & 30.0 \\                                                      % & -              
                                                         & \cmark & 9.6 & 19.2 & 40.8 & 30.4 & 33.5 \\ \midrule                                              % & 7.4            
    \multirow{2}{*}{ProxyCLIP~~\citep{lan2024proxyclip}} & \xmark & 8.4 & 20.2 & 39.1 & 26.5 & 38.1 \\                                                      % & 8.0            
                                                         & \cmark & 9.9 & 23.2 & 44.6 & 34.2 & 42.4 \\           
    \bottomrule
    \end{tabular}%
    % }
    \caption{
        Performance improvements achieved when combined with several unsupervised open-vocabulary semantic segmentation methods that do not rely on a mask generator.
        ``Load'' means whether to load model weights from \Ours{}. 
        OpenAI CLIP ViT-B/16 are used here.
    }
    \label{tab:inference_wo_mask}
\end{table}

\myPara{Additional computation beyond mask generators.}
\cref{tab:flops} presents the computational costs of different modules in our model from \cref{tab:ovss} under varying numbers of proposal masks. 
The CLIP-V adopts the ViT-L/14 architecture with an input resolution of $392^2$, 
while the mask generator employs the ConvNeXt-L architecture operating at $800^2$ resolution. 
Using 150 object categories, the table reveals that 
the primary computational bottleneck remains in the mask generator since CLIP-V for mask classification typically requires lower resolutions. 
Furthermore, the Extractor and mask generator can execute computations in parallel during implementation, thereby reducing overall processing time.

\begin{table}[t]
    \centering
    \setlength{\abovecaptionskip}{2pt}
    \setlength{\tabcolsep}{7pt}
    \small
    % \resizebox{\linewidth}{!}{%
    \begin{tabular}{cccccc}
    \toprule
    \textbf{Q}   & \multicolumn{2}{c}{\textbf{CLIP-V}} & \textbf{PSM} & \multicolumn{2}{c}{\textbf{Mask Generator}} \\
    \cmidrule(lr){1-1} \cmidrule(lr){2-3} \cmidrule(lr){4-4} \cmidrule(lr){5-6}
        & Extractor          & Fuser         &     & Backbone         & Decoder         \\
    \midrule
    100 & 416.26             & 67.05         & 0.05& 877.63           & 232.12          \\
    250 & 416.26             & 78.62         & 0.12& 877.63           & 302.35          \\
    \bottomrule
    \end{tabular}%
    % }
    \caption{FLOPs~(G) of each module.}
    \label{tab:flops}
\end{table}

% \myPara{Limitations and future work.}
% \Ours{} exhibits three primary limitations that warrant future investigation:
% \begin{itemize}
%     \item 
%     %As shown in \cref{tab:maskacc}, while our method effectively enhances CLIP's local recognition across multiple domains, 
%     %its performance remains constrained by CLIP's original recognition boundaries.
%     \Ours{} fails to distinguish new categories beyond CLIP's pretrained capability.
%     For fine-grained categories in specific domains (e.g. CUB-200~\cite{wah2011caltech}), models fine-tuned on COCO underperform original CLIP.
%     \item The mask generator's potential revealed in \cref{subfig:oracle_cls_results} stems from its redundant proposals containing texture/scale-similar out-of-vocabulary objects. 
%     Though we focus on local recognition enhancement through existing mask-based OVS inference paradigms, fundamental limitations persist: 
%     The COCO-trained mask generator usually fail on multi-granularity segmentation. So it is still worth exploring improvements in mask generation.
%     \item Current fine-tuning relies on closed-vocabulary segmentation data. Future work should explore contrastive learning with image-mask-caption triplets, enabling scalibility on fine-tuning datas.
% \end{itemize}

\section{Conclusion}
This paper addresses key challenges in open-vocabulary segmentation (OVS) by proposing \Ours{}, 
a high-quality mask tuning framework for CLIP that significantly simplifies the training paradigm. 
Our method achieves notable improvements in mask classification accuracy across multiple domain datasets, 
and when integrated with existing mask generators, demonstrates superior open-vocabulary image segmentation performances.

%% file: sec_arxiv/6_supplementary.tex
\clearpage
\appendix

\begin{figure*}[t!]
    \centering
    \begin{minipage}{0.9\textwidth}
      \includegraphics[width=\textwidth]{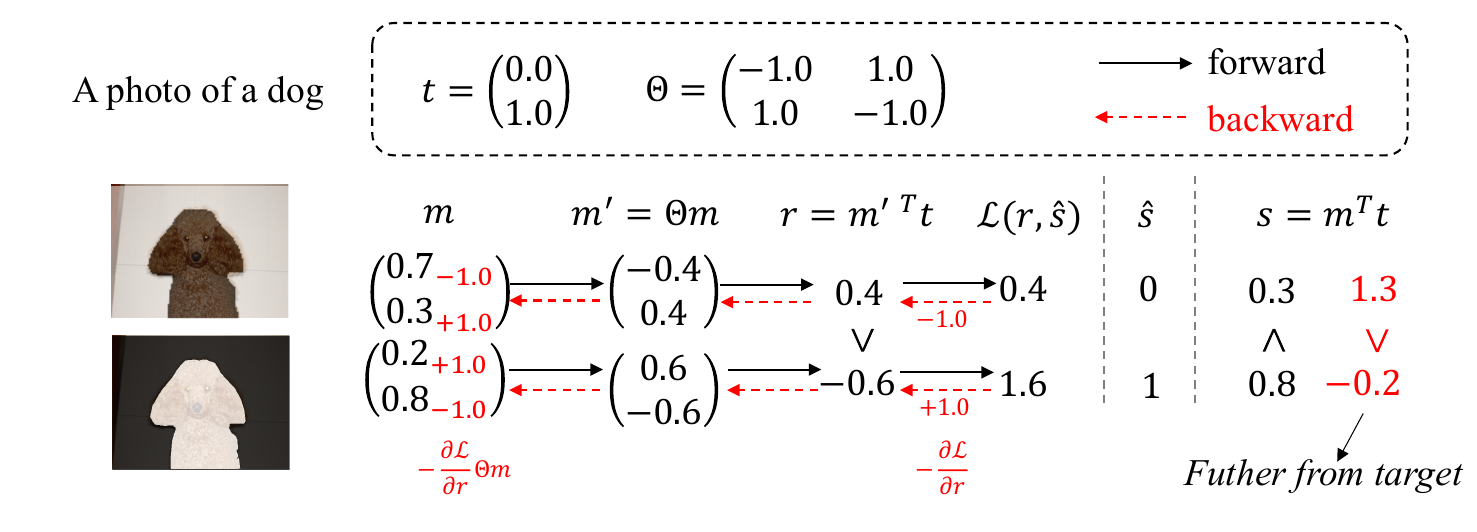}
      \caption{
          A toy example: Because the inconsistency alignment between $s$ (similarity before $\Theta$) and $r$ (similarity after $\Theta$),
          after a round of gradient descent, $s$ is further from the target $\hat{s}$.
      }
      \label{fig:pre_aligned_demo}
    \end{minipage}
\end{figure*}

% Add appendix title
\section*{Supplementary Material for ``High-Quality Mask Tuning Matters for Open-Vocabulary Segmentation''}

% In the supplementary materials, we will provide:
% \begin{itemize}
%     \item Extended methodological and implementation details. (\cref{sec:method_details})
%     \item Comprehensive dataset statistics. (\cref{sec:dataset_details})
%     \item Additional experiment results and ablations. (\cref{sec:additional_experiments})
%     \item More visualizations. (\cref{fig:vis_select})
% \end{itemize}

\section{Method Details}
\label{sec:method_details}

\subsection{Extracting mask embeddings of different architectures}
The main text exclusively details mask embedding extraction from ViT-architecture CLIP models. 
This methodology can be extended to other CLIP architectures through the following adaptations:
For architectures containing attention pooling layers without Transformer blocks, the Fuser module operates starting from the attention pooling layer, with weight functions identical to \cref{eq:phi}.
In models employing global average pooling without Transformer components, the Fuser initiates computation from the global pooling layer, where the weight function degenerates to L1 normalization to implement masked average pooling.

\subsection{Phenomenon of inconsistency alignment}\label{sec:phenomenon}
We use a toy example to illustrate the potential consequences of destroying alignment consistency.
Assume that we use the parameter $\Theta$ to update the mask embeddings, 
changing the original similarity from $s$ to $r$, and then optimize towards the target.
However, due to the lack of explicit constraints on $\Theta$,
there may be a situation as shown in \cref{fig:pre_aligned_demo}, where the order of $s$ and $r$ is inconsistent,
resulting in correct optimization on $r$ but incorrect optimization on $s$.
For simplicity,  $\mathcal{L}$ in \cref{fig:pre_aligned_demo} denotes an L1 loss without softmax, 
and only mask embeddings $m$ are updated in one round of gradient descent.

\subsection{Mask classification score integration}
The mask generator demonstrates inherent capability to produce classifications within its training vocabulary. 
When unambiguous semantic correspondence exists during inference, following established protocols in mask-based open-vocabulary segmentation (OVS) works~\cite{xu2022zsseg,xu2023odise,yu2024fcclip}, 
our method's output class scores can be adaptively aggregated across the mask generator's training vocabulary. 
This enables category probability estimation while maintaining compatibility with existing OVS frameworks.

Specifically, let $\mathcal{K}$ be the training vocabularies of the mask generator. $P_{c}$ and $P_{s}$ are the class probabilities of /Ours{} and the mask generator, respectively.
The integrated class probabilities $P_\gamma$ can be expressed as follows:
\begin{equation}\label{eq:ensemble}
    % 2 cases
    P_\gamma^{(i)} =
    \begin{cases}
      {P_{s}^{(i)}}^{\gamma} \cdot {P_{c}^{(i)}}^{(1 - \gamma)}, & \text{if} ~ i \in \mathcal{K} \\
      P_{c}^{(i)}. & \text{otherwise} \\
    \end{cases}
\end{equation}
We set $\gamma=0.1$ for our model when integrating with the mask generator.
However, as demonstrated in \cref{tab:ensemble}, 
our's performance enhancement stems primarily from improved mask classification capability rather than this integration strategy.

\section{Dataset Details}
\label{sec:dataset_details}

\begin{table}[t]
    \centering
    \small
    \setlength{\abovecaptionskip}{2pt}
    \setlength{\tabcolsep}{8pt}
    \begin{tabular}{lcr}
    \toprule
    \textbf{Dataset}                 & \textbf{\#Classes}           & \textbf{\#Images}\\
    \midrule
    \multicolumn{3}{l}{\emph{\textbf{Train}}} \\
    COCO Stuff~\cite{caesar2018cocostuff}              & 171                 & 118280  \\
    COCO Panoptic~\cite{lin2014mscoco}           & 133                 & 118287  \\
    \midrule
    \multicolumn{3}{l}{\emph{\textbf{Eval}}}  \\
    COCO Stuff              & 171                 & 5000    \\
    COCO Panoptic           & 133                 & 5000    \\
    ADE20K~\cite{zhou2017ade20k}                  & 150 / 847           & 2000    \\
    PASCAL Context~\cite{mottaghi2014context}          & 59 / 459            & 5000    \\
    PASCAL VOC~\cite{everingham2010pascal}              & 20 (w/o background) & 1449    \\
    Cityscapes~\cite{cordts2016cityscapes}              & 19                  & 500     \\
    \midrule
    \multicolumn{3}{l}{\emph{\textbf{Eval (MESS - General)}}} \\
    BDD100K~\cite{yu2020bdd100k}                 & 19                  & 1000    \\
    Dark Zurich~\cite{sakaridis2019darkz}             & 20                  & 50      \\
    MHP v1~\cite{li2017mhp}                  & 19                  & 980     \\
    FoodSeg103~\cite{wu2021foodseg}              & 104                 & 2135    \\
    ATLANTIS~\cite{erfani2022atlantis}                & 56                  & 1295    \\
    DRAM~\cite{cohen2022dram}                    & 12                  & 718     \\
    \midrule
    \multicolumn{3}{l}{\emph{\textbf{Eval (MESS - Earth)}}} \\
    iSAID~\cite{waqas2019isaid}                   & 16                  & 4055    \\
    ISPRS Potsdam~\cite{swissphoto2012isprs}           & 6                   & 504     \\
    WorldFloods~\cite{istomina2016floods}             & 3                   & 160     \\
    FloodNet~\cite{rahnemoonfar2021floodnet}                & 10                  & 5571     \\     
    UAVid~\cite{lyu2020uavid}                   & 8                   & 840     \\
    \midrule
    \multicolumn{3}{l}{\emph{\textbf{Eval (MESS - Medical)}}} \\
    Kvasir-Inst.~\cite{pogorelov2017kvasir}            & 2                   & 118     \\
    CHASE DB1~\cite{fraz2012chasedb1}               & 2                   & 20      \\
    CryoNuSeg~\cite{mahbod2021cryonuseg}               & 2                   & 30      \\
    PAXRay-4~\cite{seibold2022paxray}                & 2                   & 180     \\
    \midrule
    \multicolumn{3}{l}{\emph{\textbf{Eval (MESS - Engineer)}}} \\
    Corrosion CS~\cite{bianchi2021corrosion}            & 4                   & 44      \\
    DeepCrack~\cite{liu2019deepcrack}               & 2                   & 237     \\
    ZeroWaste-f~\cite{bashkirova2022zerowaste}             & 5                   & 929     \\
    PST900~\cite{shivakumar2020pst900}                  & 5                   & 288     \\
    \midrule
    \multicolumn{3}{l}{\emph{\textbf{Eval (MESS - Agriculture)}}} \\
    SUIM~\cite{islam2020suim}                    & 8                   & 110     \\
    CUB-200~\cite{wah2011cub200}                 & 201                 & 5794    \\
    CWFID~\cite{haug2015cwfid}                   & 3                   & 21      \\
    \bottomrule
    \end{tabular}
    \caption{Statistics of datasets used in our experiments.}
    \label{tab:datasets}
\end{table}

% \subsection{Dataset Statistics}
\cref{tab:datasets} provides detailed statistics for the datasets used in our experiments.
\cref{tab:ade_split} lists the categories in ADE20K that are present in COCO Panoptic (seen) or absent from COCO Panoptic (unseen). This split is used in our experiments to evaluate the generalization capability in \cref{sec:ablation}.

\begin{table*}[t]
    \small
    \centering
    % \tablestyle{6pt}{1.}
    
  %\vskip 0.15in
    % \begin{center}
    % \begin{small}
    % \begin{sc}
    \begin{tabular}{lc}
    \toprule
    Split & Name\\    
    \midrule
    \multirow{6}{*}{Seen~(64)}  & \emph{wall, building, sky, floor, tree, ceiling, road, bed, windowpane, grass, cabinet, sidewalk,}\\
    &\emph{person, door, table, mountain, plant, curtain, chair, car, water, sofa, shelf, sea,}\\
    &\emph{mirror, rug, fence, rock, lamp, counter, sand, sink, refrigerator, stairs, pillow, river,} \\
    &\emph{bridge, toilet, flower, book, bench, palm tree, boat, bus, towel, light bulb, truck, } \\
    &\emph{television receiver, airplane, apparel, bottle, tent, oven, food, microwave, plant pots,} \\
    &\emph{animal, bicycle, blanket, vase, traffic light, plate, cup, clock} \\
    \midrule
    \multirow{10}{*}{Unseen~(86)} & \emph{earth, painting, house exterior, field, armchair, seat, desk, wardrobe,}\\
    &\emph{bathtub, railing, cushion, pedestal, box, column, signboard, chest of drawers, skyscraper, }\\
    &\emph{fireplace, grandstand, path, runway, case, pool table, screen door, stairway, bookcase,}\\
    &\emph{window screen, coffee table, hill, countertop, stove, kitchen island, computer, swivel chair,}\\
    &\emph{arcade machine, hovel, tower, chandelier, awning, streetlight, booth, dirt track, pole, land,}\\
    &\emph{bannister, escalator, ottoman, buffet, poster, stage, van, ship, fountain, conveyer belt, }\\
    &\emph{washer, plaything, swimming pool, stool, barrel, basket, waterfall, bag, minibike, cradle, }\\
    &\emph{step, tank, trade name, lake, dishwasher, projection screen, sculpture, exhaust hood, sconce,  }\\
    &\emph{tray, ashcan, ceiling fan, pier, crt screen, monitor, bulletin board, shower, radiator,}\\
    &\emph{canopy, flag, bar, ball }\\
   \bottomrule
    \end{tabular}
    % \end{sc}
    % \end{small}
    % \end{center}
    % \vskip -0.1in
    \caption{
      Categories in ADE20K that are present in COCO Panoptic (seen) or are absent from COCO Panoptic (unseen).
    }
    \label{tab:ade_split}
\end{table*}

\begin{table*}[t]
    \small
    \setlength{\abovecaptionskip}{2pt}
    \begin{subtable}[t]{\textwidth}
        \centering
        \setlength{\tabcolsep}{15pt}
        \begin{tabular}{ccccccc}
            \toprule
            \textbf{Method} & \textbf{BDD100K} & \textbf{Dark Zurich} & \textbf{MHPv1} & \textbf{FoodSeg103} & \textbf{ATLANTIS} & \textbf{DRAM} \\
            \midrule
            OpenAI CLIP & 44.6 & 27.5 & 48.6 & 57.1 & 58.1 & 89.3 \\
            EVA02 CLIP & 44.7 & 34.3 & 49.2 & 61.4 & 61.0 & 91.0 \\
            CLIPSelf & 51.8 & 30.6 & 61.0 & 47.4 & 59.1 & 65.3 \\
            CAT-Seg  & 59.3 & 40.3 & 49.8 & 54.6 & 65.3 & 94.6 \\
            MaskCLIP++ & \textbf{67.2} & \textbf{48.3} & \textbf{64.9} & \textbf{64.7} & \textbf{64.5} & \textbf{98.0} \\
            \bottomrule
        \end{tabular}
        \caption{General}
        \label{tab:general}
    \end{subtable}

    \vspace{0.5em}

    \begin{subtable}[t]{0.64\textwidth}
        % \centering
        \raggedright
        \setlength{\tabcolsep}{9pt}
        \begin{tabular}{cccccc}
            \toprule
            \textbf{Method} & \textbf{iSAID} & \textbf{ISPRS Potsdam} & \textbf{WorldFloods} & \textbf{FloodNet} & \textbf{UAVid} \\
            \midrule
            OpenAI CLIP & 59.2 & 65.5 & 65.9 & 65.9 & 48.9 \\
            EVA02 CLIP & 61.0 & 68.4 & 71.4 & 59.6 & 42.2 \\
            CLIPSelf & 43.6 & 41.7 & 42.4 & 38.3 & 62.6 \\
            CAT-Seg  & 59.8 & 76.9 & 71.0 & 62.3 & 64.4 \\
            MaskCLIP++ & \textbf{60.5} & \textbf{80.0} & \textbf{75.1} & \textbf{63.0} & 64.4 \\
            \bottomrule
        \end{tabular}
        \caption{Earth Monitoring}
        \label{tab:earth_monitoring}
    \end{subtable}
    \hfill
    \begin{subtable}[t]{0.34\textwidth}
        % \centering
        \raggedleft
        \setlength{\tabcolsep}{8pt}
        \begin{tabular}{ccc}
            \toprule
            \textbf{SUIM} & \textbf{CUB-200} & \textbf{CWFID} \\
            \midrule
            55.5 & 54.7 & 42.8 \\
            57.6 & \textbf{60.0} & 39.7 \\
            56.0 & 51.1 & 61.9 \\
            72.9 & 47.2 & 73.0 \\
            \textbf{73.2} & 40.8 & \textbf{82.5} \\
            \bottomrule
        \end{tabular}
        \caption{Agriculture and Biology}
        \label{tab:agriculture_biology}
    \end{subtable}
    
    \vspace{0.5em}

    \begin{subtable}[t]{0.54\textwidth}
        % \centering
        \raggedright
        \setlength{\tabcolsep}{5pt}
        \begin{tabular}{ccccc}
            \toprule
            \textbf{Method} & \textbf{Kvasir Inst.} & \textbf{CHASE DB1} & \textbf{CryoNuSeg} & \textbf{PAXRay-4} \\
            \midrule
            OpenAI CLIP & 93.2 & 50.0 & 50.0 & 50.8 \\
            EVA02 CLIP & 95.0 & 50.0 & 50.0 & 51.7 \\
            CLIPSelf & 72.0 & 50.0 & 50.0 & 50.0 \\
            CAT-Seg  & 89.8 & 50.0 & 50.0 & 58.1 \\
            MaskCLIP++ & \textbf{99.2} & 50.0 & 50.0 & \textbf{99.4} \\
            \bottomrule
        \end{tabular}
        \caption{Medical Sciences}
        \label{tab:medical_sciences}
    \end{subtable}
    \hfill
    \begin{subtable}[t]{0.44\textwidth}
        % \centering
        \raggedleft
        \setlength{\tabcolsep}{4pt}
        \begin{tabular}{ccccc}
            \toprule
            \textbf{Corrosion CS} & \textbf{DeepCrack} & \textbf{ZeroWaste-f} & \textbf{PST900} \\
            \midrule
            25.5 & 69.2 & 32.5 & 19.4 \\
            25.0 & 58.9 & 33.0 & 18.4 \\
            \textbf{31.6} & 76.8 & 45.2 & 27.5 \\
            21.6 & 61.0 & 30.2 & 26.7 \\
            24.4 & \textbf{85.7} & \textbf{57.4} & \textbf{33.8} \\
            \bottomrule
        \end{tabular}
        \caption{Engineering}
        \label{tab:engineering}
    \end{subtable}
    \vspace{0.5em}
    \caption{Mask accuracy on MESS datasets~\cite{MESSBenchmark2023}.}
    \label{tab:mess}
\end{table*}

\section{Additional Experiments}
\label{sec:additional_experiments}

\subsection{Mask accuracy on MESS}
\cref{tab:mess} complements \cref{tab:maskacc} by listing the mask accuracy of different models on each dataset in MESS~\cite{MESSBenchmark2023}.
\Ours{} achieves better maskAcc on 17 out of 22 datasets, 
providing compelling empirical evidence that the proposed method 
significantly enhances mask classification accuracy while simultaneously preserving the inherent generalization capabilities of the original CLIP.

However, note that in specialized fine-grained classification tasks like the bird-oriented CUB-200 dataset~\cite{wah2011cub200}, 
where category labels are semantically close and images usually have a single foreground subject, 
fine-grained category domain-specific knowledge impacts performance more than localized perception. 
Thus, CLIP models fine-tuned on COCO generally perform worse than the original CLIP in such cases.

\subsection{Effectiveness of mask tuning on different CLIP architectures}
To validate the generalizability of our approach across varying CLIP architectures, 
we conduct fine-tuning experiments on the COCO Panoptic dataset~\cite{lin2014mscoco} using distinct CLIP architectures
and report the mask classification accuracy of our method as documented in \cref{tab:arch}.

\begin{table}[t]
    \small
    \centering
    \setlength{\abovecaptionskip}{2pt}
    \setlength{\tabcolsep}{4pt}
    \begin{tabular}{cccccc}
    \toprule
    \textbf{Arch.} & \textbf{A-847} & \textbf{PC-459} & \textbf{A-150} & \textbf{PC-59} & \textbf{Stuff} \\
    \midrule
    RN50x16      & 24.1 & 37.5 & 50.4 & 73.3 & 54.1 \\
    ConvNeXt-B   & 30.0 & 42.1 & 56.4 & 76.3 & 55.6 \\
    ConvNeXt-L   & 33.0 & 50.8 & 61.0 & 78.7 & 59.0 \\
    ConvNeXt-XXL & 38.7 & 56.1 & 64.6 & 81.5 & 62.5 \\
    ViT-B/16     & 29.3 & 45.7 & 58.3 & 78.6 & 58.0 \\
    ViT-L/14     & 37.5 & 56.9 & 66.0 & 83.2 & 63.6 \\
    \bottomrule
    \end{tabular}
    \caption{
        Mask accuracy of \Ours{} on different architectures fine-tuned on COCO Panoptic~\cite{lin2014mscoco}.
    }
    \label{tab:arch}
\end{table}

\subsection{Other ablation studies}
\myPara{Effect of Fuser depth.}
% 不同的掩码插入位置会对模型性能产生一定的影响。
% 根据\cref{sec:framework}，Fuser 包含的 Transformer blocks 数量可以表示为 $L - K$。
% \cref{tab:fuser_depth}展示了不同的 Fuser 深度对 OVS 性能的影响。
% 考虑到 Fuser 深度增加会增加与掩码数量成正比的计算量，我们选择 $L - K = 2$ 作为默认设置。
The positioning of mask insertions exerts non-negligible effects on model performance. 
As delineated in \cref{sec:framework}, the quantity of Transformer blocks in the Fuser module can be mathematically formulated as $L - K$. 
\cref{tab:fuser_depth} presents a systematic analysis of how varying Fuser depths impact OVS performance. 
Considering the computational overhead that scales linearly with mask quantity when increasing Fuser depth, we empirically select $L - K = 2$ as the default configuration.
\begin{table}[h]
    % \centering
    \small
    \setlength{\abovecaptionskip}{2pt}
    \begin{subtable}[t]{0.5\linewidth}
        \raggedright
        \setlength{\tabcolsep}{10pt}
        \begin{tabular}{ccc}
        \toprule
        $\bm{L - K}$ & \textbf{mIoU} &  \textbf{PQ}  \\ 
        \midrule
        1                  &     33.3      &     24.1      \\
        2                  &     33.5      & \textbf{24.5} \\
        3                  & \textbf{33.8} &     24.4      \\
        4                  &     33.5      &     24.4      \\ 
        \bottomrule
        \end{tabular}%
        \caption{Effect of Fuser depth.}
        \label{tab:fuser_depth}
    \end{subtable}
    \hfill
    \begin{subtable}[t]{0.45\linewidth}
        \raggedleft
        \setlength{\tabcolsep}{10pt}
        \begin{tabular}{cc}
        \toprule
        $\bm{\alpha}$ & \textbf{mIoU} \\
        \midrule
        $e^{5}$     & 33.1 \\
        $e^{-5}$    & 33.5 \\
        learnable & \textbf{33.8} \\
        \bottomrule
        \end{tabular}
        \caption{Effect of $\alpha$.}
        \label{tab:alpha}
    \end{subtable}
    \caption{Ablation studies on the effect of Fuser depth and $\alpha$.}
\end{table}

\myPara{Effect of $\alpha$.}
As defined in \cref{eq:phi}, the bigger $\alpha$ is, the less the semantic similarity of the image context is utilized.
As $\alpha \rightarrow +\infty$, $E_m^{(l+1)}$ degenerates to the mask average pooling result of $F^{(l)}$, thereby inadequately leveraging contextual information beyond the masked regions. 
Conversely, when $\alpha=0$, $E_m^{(l+1)}$ reduces to the CLS token after $l$-th layer.
% 如表所示，$\alpha$ 选择较小的值具有更好地性能，因此我们最终将 $\alpha$ 初始化为 $e^{-5}$ 并在训练过程中在对数空间中调整。
As shown in \cref{tab:alpha}, selecting a smaller $\alpha$ value yields superior performance.
Therefore, we initialize $\alpha$ as $e^{-5}$ and adaptively adjust it in the logarithmic space during training.

\myPara{Effect of classification ensemble.}
% 为了证明我们的方法的性能提升主要来自于改进的掩码分类能力，我们在\cref{tab:ensemble}中列出了是否使用集成策略带来的影响。
To demonstrate that the performance gains of our method primarily stem from enhanced mask classification capabilities, 
we present in \cref{tab:ensemble} the comparative analysis of implementing versus omitting the ensemble strategy.

\begin{table}
    \centering
    \small
    \setlength{\abovecaptionskip}{2pt}
    \setlength{\tabcolsep}{4pt}
    \begin{tabular}{lccccc}
        \toprule
        \textbf{Method} & \textbf{Ensemble} & \textbf{A-847} & \textbf{PC-459} & \textbf{A-150} & \textbf{PC-59} \\ %& \textbf{PAS-20} \\
        \midrule
        \multirow{2}{*}{FC-CLIP}            & \xmark & 12.1 & 12.7 & 27.4 & 42.8 \\ %&  \\
                                            & \cmark & 14.8 & 18.2 & 34.1 & 58.4 \\ %& 95.4 \\
        \multirow{2}{*}{\Ours{}}            & \xmark & 15.3 & 21.4 & 36.8 & 62.4    \\ %& 96.4 \\
                                            & \cmark & 15.4 & 21.3 & 37.1 & 62.6 \\ %& 96.4 \\
        \bottomrule
    \end{tabular}
    \caption{Effect of classification ensemble.}
    \label{tab:ensemble}
\end{table}

% \section{Additional Visualizations}
% \cref{fig:vis_select} provides more visualizations of semantic- and instance-level open-vocabulary segmentation.
% Images are from ADE20K~\citep{zhou2017ade20k}.
% The confidience threshold for instance segmentation is 0.3.

\newcommand{\addFigApp}[1]{
  \begin{minipage}[t]{\textwidth}
    \centering
    \includegraphics[width=0.245\textwidth]{figures/vis_select/img/#1.jpg}
    \hfill
    \includegraphics[width=0.245\textwidth]{figures/vis_select/sem_seg/#1.jpg}
    \hfill
    \includegraphics[width=0.245\textwidth]{figures/vis_select/pan_seg/#1.jpg}
    \hfill
    \includegraphics[width=0.245\textwidth]{figures/vis_select/ins_seg/#1.jpg}
    \hfill
  \end{minipage}
}

\begin{figure*}
  \small
  \begin{minipage}[t]{\textwidth}
    \hspace{0.085\textwidth}
    Image 
    \hspace{0.140\textwidth}
    Semantic Segmentation
    \hspace{0.08\textwidth}
    Panoptic Segmentation
    \hfill
    Instance Segmentation
    \hspace{0.04\textwidth}
  \end{minipage}
  \vspace{-0.8em}
  \addFigApp{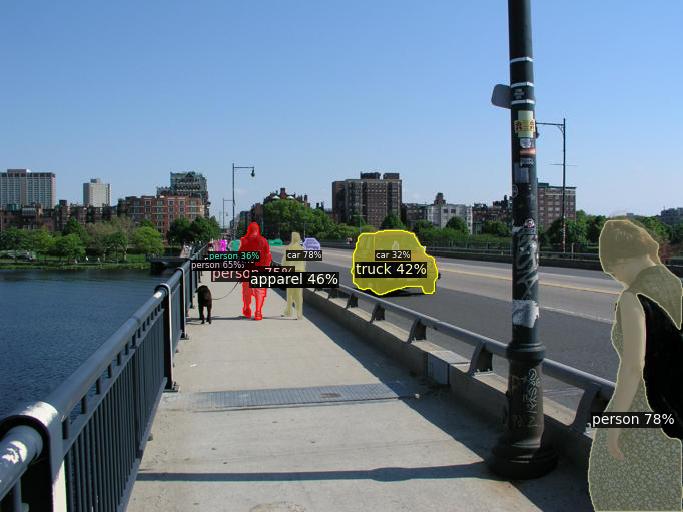}
  \vspace{-0.8em}
  \addFigApp{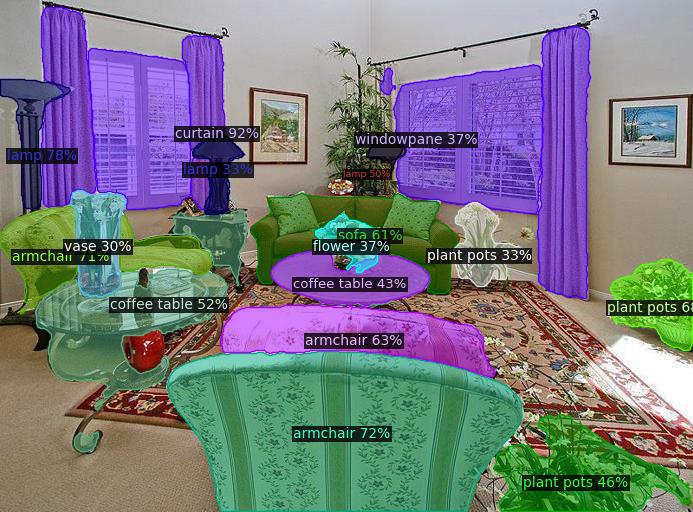}
  \vspace{-0.8em}
  \addFigApp{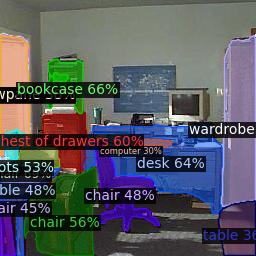}
  \vspace{-0.8em}
  \addFigApp{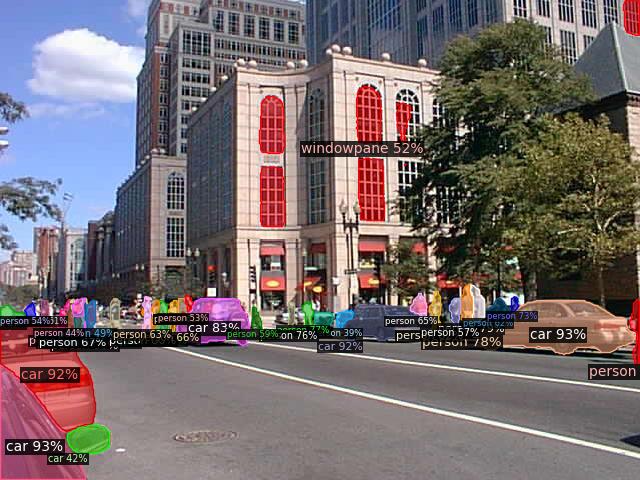}
  % \addFigApp{ADE_val_00001233}
  \vspace{-0.8em}
  \addFigApp{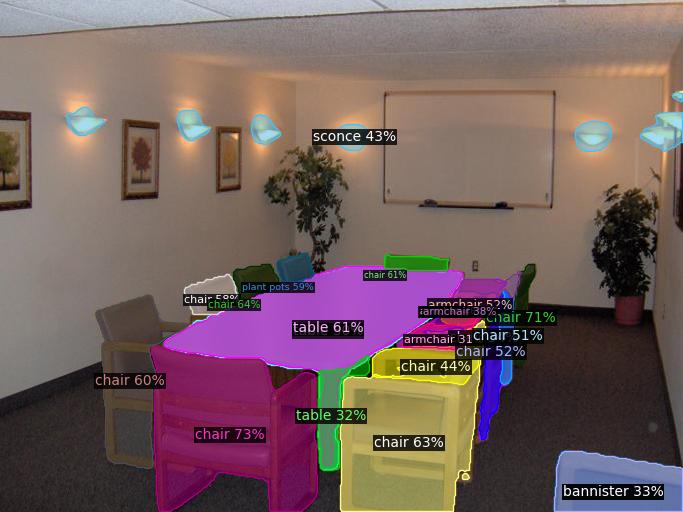}
  \vspace{-0.8em}
  \addFigApp{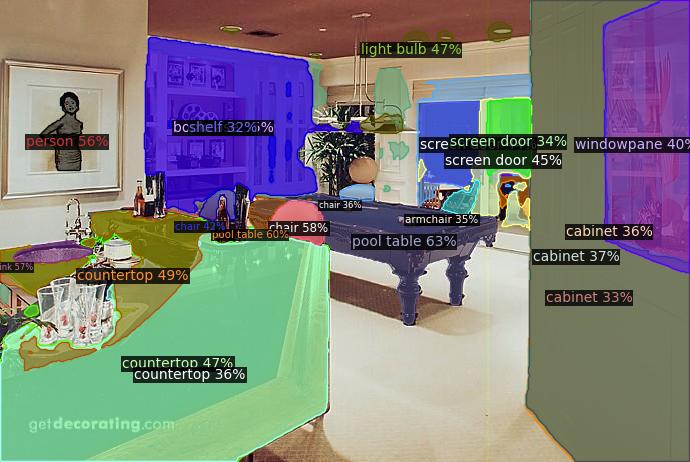}
  \vspace{-0.8em}
  % \addFigApp{ADE_val_00001421}
%   \addFigApp{ADE_val_00001480}
  % \addFigApp{ADE_val_00001534}
  
  % \addFigApp{ADE_val_00001715}
%   \vspace{-1em}
  \caption{
    More visualizations of semantic- and instance-level of open-vocabulary segmentation.
    We use mask generator from FC-CLIP~\citep{yu2024fcclip} with ConvNeXt-L backbone.
    And use CLIP ViT-L/14 fine-tuned on COCO Panoptic~\citep{lin2014mscoco}. 
    Images are from ADE20K~\citep{zhou2017ade20k}.
    The confidience threshold for instance segmentation is 0.3.
  }
  \label{fig:vis_select}
\end{figure*}